\documentclass{article}

\PassOptionsToPackage{numbers, compress}{natbib}
\usepackage[numbers]{natbib}

\usepackage{graphicx}

\usepackage[preprint]{neurips_2024}

\usepackage[utf8]{inputenc} % allow utf-8 input
\usepackage[T1]{fontenc}    % use 8-bit T1 fonts
\usepackage{hyperref}       % hyperlinks
\usepackage{url}            % simple URL typesetting
\usepackage{booktabs}       % professional-quality tables
\usepackage{amsfonts}       % blackboard math symbols
\usepackage{nicefrac}       % compact symbols for 1/2, etc.
\usepackage{microtype}      % microtypography
\usepackage{xcolor}         % colors
\usepackage{amsmath}

\title{\textbf{QIANets: Quantum-Integrated Adaptive Networks for Reduced Latency and Improved Inference Times in CNN Models}
}

\author{%
Zhumazhan Balapanov \quad Vanessa Matvei \quad Olivia Holmberg\quad Edward Magongo \\ \quad \textbf{Jonathan Pei} \quad \textbf{Kevin Zhu}
\\
Algoverse AI Research\\
\texttt{kevin@algoverse.us}
}

\begin{document}

\maketitle

\begin{abstract}
Convolutional neural networks (CNNs) have made significant advances in computer vision tasks, yet their high inference times and latency often limit real-world applicability. While model compression techniques have gained popularity as solutions, they often overlook the \textit{critical balance between low latency and uncompromised accuracy.} By harnessing \textbf{quantum-inspired pruning},\textbf{ tensor decomposition} and \textbf{annealing-based matrix factorization} – three quantum-inspired concepts – we introduce QIANets: a novel approach of redesigning the traditional GoogLeNet, DenseNet, and ResNet-18 model architectures to process more parameters and computations whilst maintaining low inference time.  Code: \href{https://github.com/edwardmagongo/Quantum-Inspired-Model-Compression}{https://github.com/quantum-inspired-model-compression}
\end{abstract}

\section{Introduction}
The field of computer vision (CV) has recently experienced a substantial rise in interest \cite{su2021affective}. This surge has created transformative advancements, driving the development of deep learning models, particularly within convolutional architecture, such as DenseNet \cite{huang2017densely}, GoogLeNet \cite{szegedy2015going}, and ResNet-18 \cite{he2016deep}. These methods have significantly optimized neural networks for image processing tasks, achieving state-of-the-art performance across multiple benchmarks \cite{anumol2023advancements}. However, the increasing computational complexity, memory consumption, and model size (millions to billions of parameters) pose substantial challenges for deployment, especially in time-sensitive and computationally-limited scenarios. The demand for \textit{low-latency processing} in real-time applications, such as image processing and automated CV systems, is critical; compact models are needed for faster responses \cite{honegger2014real}.

To address these issues, researchers have explored various optimization techniques to reduce inference times and latency while maintaining high accuracy. Model compression techniques such as pruning, quantization, and knowledge distillation have shown promise in enhancing model efficiency \cite{li2023model}. Yet, these methods often come with trade-offs that can impact model performance, necessitating a careful balance between energy efficiency and accuracy.

In recent years, the principles of quantum computing have emerged as an avenue for accelerating inference in machine learning \cite{divya2021quantum}. Quantum-inspired methods, which leverage phenomena such as quantum optimization algorithms, strive to maintain model performance by reducing computational requirements, thereby offering significant speedups for certain tasks \cite{pandey2023quantum}. Meanwhile, traditional model compression techniques reduce the size of neural networks by removing less important weights, \textit{sacrificing accuracy for lower latency} \cite{francy2024edge}. By integrating concepts from quantum mechanics into convolutional neural network (CNN) models, our approach seeks to address these limitations. We explore the potential of designing CNNs to balance improved inference times with minimal accuracy loss, creating a novel solution.

Within this context, we employ three key quantum-inspired principles: 1. quantum-inspired pruning: reducing model size by removing unnecessary parameters, guided by quantum approximation algorithms; 2. tensor decomposition: breaking down high-dimensional tensors into smaller components to reduce computational complexity; and 3. annealing-based matrix factorization: optimizing matrix factorization by using annealing techniques to find efficient representations of the data.

Our work addresses the following research question: \textbf{How can principles from quantum computing be used to design and optimize CNNs to reduce latency and improve inference times, while still maintaining stable accuracies across various models?}

In this paper, we propose a Quantum-Integrated Adaptive Networks (QIANets) – a comprehensive framework that \textit{integrates} these quantum computing techniques into the DenseNet, GoogLeNet, and ResNet-18 architectures. To the best of our knowledge, this is the first attempt made to: 1) apply quantum computing-inspired algorithms into the models’ architectures to reduce computational requirements and achieve efficient performance improvements\textit{, }and 2) specifically target these models.

The contributions of this work include:

\begin{itemize}
    \item QIANets: a comprehensive framework that integrates QAOA-inspired pruning, tensor decomposition and quantum annealing-inspired matrix factorization into three CNNs.
    \item An exploration of the trade-offs between latency, inference time, and accuracy, highlighting the effects of applying quantum principles to CNN models for real-time optimization.
\end{itemize}

\section{Related Works}
\label{gen_inst}
Our proposed method builds upon the ideas of model compression \& quantum-inspired techniques to improve the inference times of CNNs.

\subsection{Model Compression Techniques: }\textit{Pruning} is one of the most effective ways to accelerate CNNs. Cheng et al. (2018) \cite{cheng2017survey} reviewed model compression techniques for deep neural networks (DNNs), focusing on parameter pruning: removing individual weights based on importance to reduce model size while generally preserving performance.

Despite the advancements in parameter pruning, overall conventional pruning techniques have limitations: 1) high cost when applied \textit{during} training and 2) the risk of prematurely removing important data. Hou et al. (2022) \cite{hou2022chex} introduced CHEX, a \textit{training-based channel pruning} and regrowing method that reallocates channels across layers using a column subset selection (CSS) formulation, achieving significant compression without a fully pre-trained model.

\subsection{Quantum-Inspired Techniques for CNNs:}
Quantum computing is currently recognized as a potential game-changer for various fields, including NLP, due to its ability to process complex data more efficiently than classical computers. 

Shi et al. (2021) \cite{shi2022quantum} proposed a quantum-inspired architecture (QICNNs) with complex-valued weights to enhance CNN representational capacity, achieving higher accuracy and faster convergence on datasets than standard CNNs. In contrast, our methodology prioritizes structural optimization for greater computational efficiency, reducing latency and improving inference times through quantum techniques.

Hu et al. (2022) \cite{hu2022quantum} set a high standard in the field by addressing quantum neural networks (QNNs) compression. Their CompVQC framework leverages an alternating direction method of multipliers (ADMM) approach, achieving a remarkable reduction in circuit depth by over 2.5× with less than 1\% accuracy loss. While their results in QNN compression are impressive, our research introduces a novel first-attempt technique that applies QAOA-inspired pruning, tensor decomposition and quantum annealing-inspired matrix factorization to \textit{classical CNNs}, potentially complementing their approach and enhancing model efficiency.

\section{Methodology}

\subsection{Quantum-Inspired Pruning}
We build upon the established technique of \textit{pruning} to reduce the complexities of CNNs, as demonstrated in early studies (\cite{lecun1989backpropagation}; \cite{hanson1988comparing}; \cite{hassibi1993optimal}). However, we introduce a new optimization way, utilizing the \textit{Quantum Approximate Optimization Algorithm} (QAOA) \cite{farhi2022quantum} to frame pruning as a probabilistic optimization problem. For a neural network layer represented by weights as a tensor:  \textit{\(W \in \mathbb{R}^{C_{out} \times C_{in} \times H \times W} \)}, we define the importance of each weight using its absolute value: 

\begin{equation}I_{i,j} = |W_{i,j}|\end{equation}

To facilitate decision-making regarding weight retention, we normalize these importance scores with the softmax function:
\begin{equation}
P_{i,j} = \frac{e^{I_{i,j}}}{\sum_{k,l} e^{I_{k,l}}}
\end{equation}

These probabilities are then used in a quantum-inspired decision-making process. Weights are pruned based on a threshold \(\lambda\), influenced by a hyperparameter known as layer sparsity \(\alpha\):

% $\alpha$:\textit{
% \[R_{i,j} = \begin{cases} 
% 1 & \text{if } P_{i,j} \geq \lambda \\
% 0 & \text{otherwise}
% \end{cases}
% \]}

% \[
% R_{i,j} = 
% \begin{cases} 
% 1 & \text{if } P_{i,j} \geq \lambda \\
% 0 & \text{otherwise}
% \end{cases}
% \]

\begin{equation}
R_{i,j} = 
\begin{cases} 
1 & \text{if } P_{i,j} \geq \lambda \\
0 & \text{otherwise}
\end{cases}
\end{equation}

Here, \(R_{i,j}\) serves as a binary retain mask indicating whether a weight is pruned (set to 0) or retained. The threshold \(\lambda\) is calibrated to ensure that approximately \(100\alpha\%\) of the weights are pruned.

When implemented, we adopt an iterative approach across multiple stages, recalculating the retain mask based on updated probabilities. To enhance this process, we introduce a neighboring entanglement mechanism: when a weight is pruned, adjacent weights in the tensor may also be pruned with the probability \(P_{\text{entangle}}\), simulating quantum entanglement and reflecting correlated behavior among nearby weights. For convolutional layers, this sequential pruning strategy is executed over several iterations, progressively reducing the number of parameters.

\subsection{Tensor Decomposition}

\textit{Tensor decomposition} further reduces the dimensionality of the weight tensor while preserving essential information for accurate predictions. Inspired by Quantum Circuit Learning (QCL), high-dimensional tensors are decomposed into lower-dimensional forms for efficient training of quantum circuits \cite{mitarai2018quantum}.

For a weight tensor \(W \in \mathbb{R}^{C_{\text{out}} \times C_{\text{in}} \times H \times W}\), we use Singular Value Decomposition (SVD) \cite{wang2021variational} to its flattened  representation \(W_f \in \mathbb{R}^{C_{\text{out}} \times (C_{\text{in}} \cdot H \cdot W)}\): 
\begin{equation}
W_f = U \Sigma V^T
\end{equation}

Here, \( U \) and \( V \) are orthogonal matrices, and \( \Sigma \) is a diagonal matrix of singular values. The rank \( r \), chosen as a hyperparameter, controls the compression by retaining only the top \( r \) singular values:

\begin{equation}
W_f \approx U_r \Sigma_r V_r^T
\end{equation} 
After decomposition, we reconstruct the original weight tensor using truncated matrices, significantly decreasing parameters without greatly affecting model performance.

\subsection{Quantum Annealing-Inspired Matrix Factorization}
Quantum annealing optimizes systems toward their lowest energy state \cite{gherardi2024analysis}. We apply this concept to factorize weight tensors, treating it as an optimization problem aimed at minimizing the difference between the original and factorized weights.

Given a weight matrix \(W \in \mathbb{R}^{m \times n}\), we factor it into two lower-dimensional matrices, \(W_1 \in \mathbb{R}^{m \times r}\) and \(W_2 \in \mathbb{R}^{r \times n}\), where \(r\) is a hyperparameter that controls the rank:
\begin{equation}
W \approx W_1 W_2
\end{equation}

The objective is to minimize the reconstruction error: \begin{equation}
L(W_1, W_2) = \| W - W_1 W_2 \|_F^2
\end{equation}

Here, \(\| \cdot \|_F\) denotes the Frobenius norm. We use an iterative optimization procedure based on quantum annealing to minimize this loss.

The factorization employs gradient-based optimization, initializing \(W_1\) and \(W_2\) randomly. Using an optimizer like LBFGS, suitable for small parameter sets and non-convex landscapes, we simulate quantum annealing by gradually reducing step size to ensure convergence to a local minimum. Once complete, the compressed weight matrix is defined as: 
\begin{equation}
W_c = W_1 W_2
\end{equation}This compressed matrix replaces the original matrix, reducing model complexity.

% \begin{figure}[t]
%     \centering
%     \begin{minipage}[b]{0.45\textwidth}  % Adjust the width of the image container
%         \centering
%         \includegraphics[scale=0.3]{Algoverse.png}  % Your image file
%     \end{minipage}
%     \hspace{0.05\textwidth}  % Horizontal space between the image and the caption
%     \begin{minipage}[b]{0.45\textwidth}  % Adjust the width of the caption container
%         \centering
%         \caption{An illustrative diagram showcasing the framework used for Quantum-Inspired Pruning, Tensor Decomposition, and Quantum Annealing-Inspired Matrix Factorization.}
%     \end{minipage}
% \end{figure}

\begin{figure}[t]
    \centering
    \includegraphics[scale=0.5]{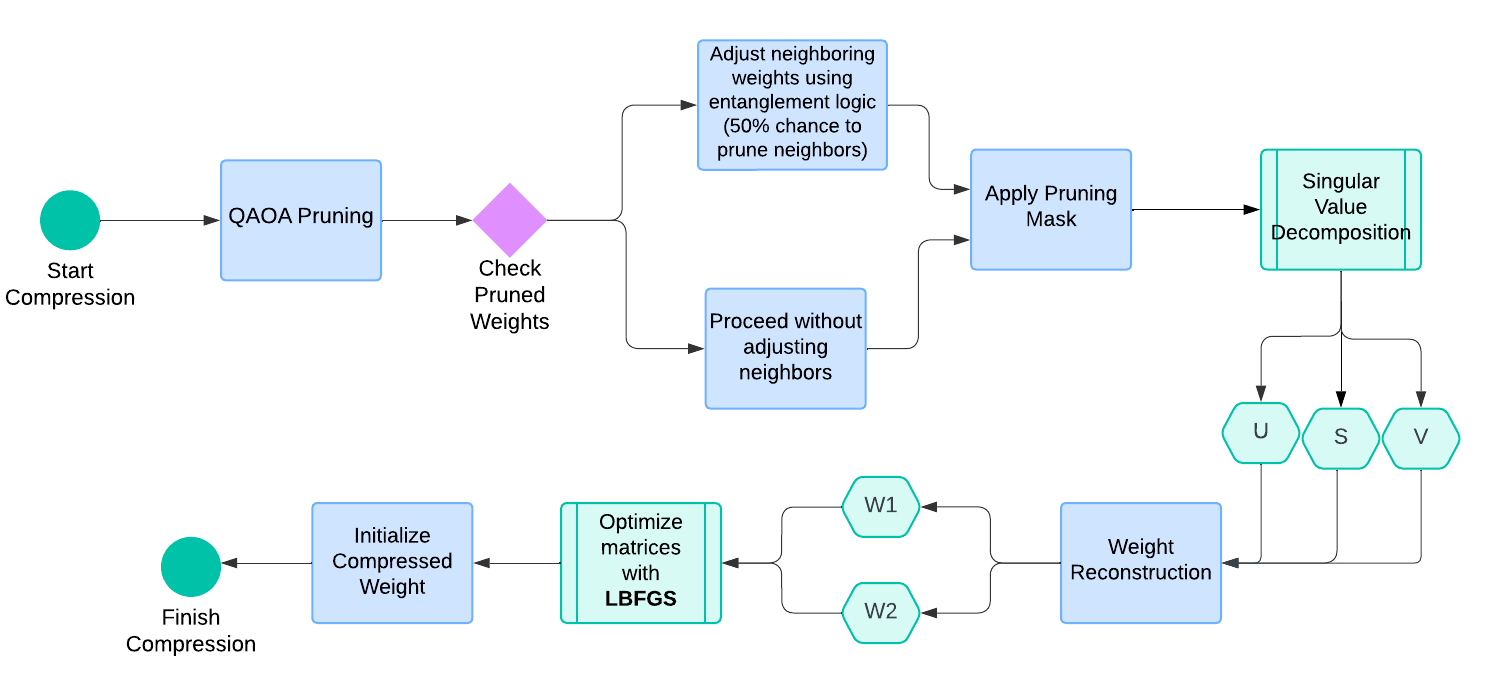}  % Your image file
    \caption{An illustrative diagram showcasing the QIANets framework}
\end{figure}

\section{Experiments and Results}
\label{others}
We applied our method to compress three CNNs: DenseNet, GoogLeNet, and ResNet-18, all on the CIFAR-10 dataset. These networks were selected for their different design structures and computational demands, such as parameter count, depth, and layer types, providing a comprehensive assessment of our method's effectiveness across different models. We evaluated the models for image classification performance, focusing on metrics such as inference time, speedup ratio, and accuracy.

Each experiment involved 1) applying the QIANets framework to the respective model architecture and 2) evaluating the models and comparing them to their baseline counterparts. \textit{The results, including networks’ changes before and after compression, are shown in Table 1.}

\begin{table}[ht]
\centering
\caption{Model Performance Comparison Before and After Compression with Rounded Figures}
\resizebox{\textwidth}{!}{%
\begin{tabular}{@{}lllllll@{}}
\toprule
\textbf{Model} & \textbf{Base Accuracy} & \textbf{Base Latency} & \textbf{New Accuracy} & \textbf{New Latency} & \textbf{Compression Ratio} \\ \midrule
\textbf{GoogleNet} & 94\% & 0.00096 T/I & 86\% & 0.00083 T/I & x1.9 \\
\textbf{ResNet} & 93\% & 0.00011 T/I & 87\% & 0.00007 T/I & x1.6 \\
\textbf{DenseNet} & 94\% & 0.000050 T/I & 88\% & 0.000042 T/I & x1.8 \\ \bottomrule
\end{tabular}%
}
\end{table}

\subsection{Experimental Setup}
The experiments were conducted within the PyTorch framework, utilizing the CIFAR-10 dataset \cite{krizhevsky2009learning}. The CIFAR-10 dataset, which consists of 60,000 32x32 RGB color images in 10 classes, with 6,000 images per class, was split into training and validation sets with an 80/20 ratio. Data preprocessing included resizing images to 224x224 pixels and normalizing pixel values to the range [-1, 1]. Moreover, to generate data variability, data augmentation strategies (random horizontal flipping and cropping) were applied, improving performance on unseen data.

All computations were accelerated using CUDA on an NVIDIA A40 GPU via Runpod. Each model underwent training for 50 epochs utilizing the Adam optimizer (learning rate = 0.001, weight decay = 1e-4), using approximately 1.6 million TFLOPS-seconds of compute. To ensure consistency across models, batch sizes of 128 were used for training, while batch sizes of 256 were employed for both evaluation and testing within the dataset. The training process involved 10 trials of 10 epochs each, followed by a final trial of 50 epochs. 

\subsection{Hyperparameter Tuning}
Hyperparameter tuning is performed using Optuna, a framework that implements various techniques to optimize certain parameters. Optuna tests various combinations of hyperparameters (batch size, learning rate and ECA Kernel Size) and dynamically adjusts them based on each trial. Following each trial, validation accuracy is calculated to test the effectiveness of current parameters. This information refines the subsequent parameters, ultimately approaching optimized parameter configurations.

\subsection{Model Specific Analysis}

To effectively accommodate the unique architectures of each model, we made minimal targeted adjustments to the QIANets method but ensured that all models were trained and evaluated under consistent and fair conditions throughout the experiments. \textit{See Appendix A for additional details}

\subsubsection{GoogLeNet}
GoogLeNet is a convolutional network with nine multi-scale processing Inception modules. The QIANets framework targets these modules, reducing the weight in their convolutions: layer sparsity of 0.1417, while employing a rank of 41 to efficiently decompose and factorize these weights. \textit{See Table 2.}

\begin{table}[h]
\centering
\caption{Training and validation metrics for GoogLeNet}
\resizebox{\textwidth}{!}{%
\begin{tabular}{@{}lll@{}}
\toprule
\textbf{Metric} & \textbf{Quantum-Inspired GoogLeNet} & \textbf{Baseline GoogLeNet} \\ \midrule
\textbf{Training Loss} & 0.7786 & 1.7066 (Epoch 1) \\
\textbf{Validation Accuracy} & 80.19\% & 38.52\% (Epoch 1) \\
\textbf{Test Loss} & 0.5732 & 0.2557 \\
\textbf{Test Accuracy} & 86.65\% & 94.29\% \\
\textbf{Average Inference Time/Image} & 0.000835 seconds (13.65\% Faster) & 0.000967 seconds \\ \bottomrule
\end{tabular}%
}
\end{table}

\subsubsection{DenseNet}
We experimented on DenseNet – a CNN structured with 12 dense blocks, with layer-by-layer connections. This intricate connectivity requires careful application of QAOA pruning to ensure that weight removal does not disrupt the model’s residual stream. \textit{See Table 3.}

\begin{table}[h]
\centering
\caption{Training and validation metrics for DenseNet}
\resizebox{\textwidth}{!}{%
\begin{tabular}{@{}lll@{}}
\toprule
\textbf{Metric} & \textbf{Quantum-Inspired DenseNet} & \textbf{Baseline DenseNet} \\ \midrule
\textbf{Training Loss} & 0.5351 & 2.3027 \\
\textbf{Validation Accuracy} & 81.33\% & 10.34\% \\
\textbf{Test Loss} & 0.4712 & 0.2462 \\
\textbf{Test Accuracy} & 88.52\% & 94.05\% \\
\textbf{Average Inference Time/Image} & 0.000042 seconds (15.20\% faster) & 0.000050 seconds \\ \bottomrule
\end{tabular}%
}
\end{table}

\subsubsection{ResNet-18}
Lastly, we tested on ResNet-18, a CNN characterized by its unique\textbf{ }residual learning framework and shortcut connections. The QIANets framework targets the residual blocks in the model, reducing less significant weights and channels, detected by ECA’s straightforward 1D convolution. \textit{See Table 4.}

\begin{table}[h]
\centering
\caption{Training and validation metrics for ResNet-18}
\resizebox{\textwidth}{!}{%
\begin{tabular}{@{}lll@{}}
\toprule
\textbf{Metric} & \textbf{Quantum-Inspired ResNet-18} & \textbf{Baseline ResNet-18} \\ \midrule
\textbf{Training Loss} & 0.4078 & 0.6501 \\
\textbf{Validation Accuracy} & 90.25\% & 91.30\% \\
\textbf{Test Loss} & 0.6447 & 0.3195 \\
\textbf{Test Accuracy} & 87.11\% & 93.11\% \\
\textbf{Average Inference Time/Image} & 0.00007 seconds (36.4\% faster) & 0.00011 seconds \\ \bottomrule
\end{tabular}%
}
\end{table}

\subsection{Analysis of the QIANets Framework}

The QIANets framework achieves compression ratios of 1.9× for GoogleNet, 1.8× for DenseNet, and 1.6× for ResNet-18, demonstrating effective latency reductions. Each model showed consistent loss reduction, approached baseline accuracy post-fine-tuning, and achieved faster inference. Although results are slightly below some CNN compression methods \cite{han2015deep}, QIANets shows promise for quantum-inspired compression.

\section{Conclusion}
In this paper, we introduced the QIANets framework, applying it to DenseNet, GoogLeNet, and ResNet-18 to reduce latency and improve inference time while preserving accuracy. Our results highlight the potential of quantum-inspired techniques for CNN compression, yielding valuable insights into the trade-offs between latency and accuracy across various architectures.

\section{Limitations}
While our results demonstrate the potential of QIANets and quantum-inspired principles in model compression, they also highlight several factors influencing performance. Future work should address these limitations through more in-depth experiments assessing the scalability and practical relevance of quantum-inspired techniques.

\begin{enumerate}
    \item \textbf{Data Constraints}: The evaluation was limited to the relatively simple CIFAR-10 dataset, which may not fully capture the diversity, complexity, or scalability challenges present in larger real-world datasets. Additionally, due to the computational expense, the approach was tested on a restricted number of trials.
    \item \textbf{Model Adaptation}: The lack of adaptation across different architectures may hinder the QIANets framework's ability to balance latency and accuracy. Performance in certain scenarios does not guarantee similar results across architectures without model-specific adjustments, complicating future adaptations.
    \item \textbf{Hardware Limitations}: This study does not address hardware-specific limitations. Our techniques have yet to be optimized for specialized hardware, such as custom FPGAs or GPUs, which could further reduce latency and improve data throughput.
\end{enumerate}

\begin{ack}
This work was completed through the Algoverse program, and we acknowledge the team for their support. We also thank Sean O'Brien and the anonymous reviewers for their valuable feedback.
\end{ack}

% Define the full reference

\bibliographystyle{unsrt}  % or another style like abbrv, alpha, etc.
\bibliography{references}

\newpage
\appendix

\section{Extended Results and Model Analysis}
\subsection{GoogLeNet:}
\textbf{Performance Progression Across Epochs: }During Trial 0 of optimization, the model began with a validation accuracy of 21.08\% at Epoch 1 and quickly progressed to 70.18\% by Epoch 10, indicating\textit{ rapid learning during the first stages of training}. The highest validation accuracy was 80.19\%\textbf{ }throughout Trial 5. The final quantum-inspired GoogLeNet’s test accuracy was 86.65\%, a notable improvement from the earlier 80.19\%, approaching the baseline accuracy of 94.29\%\textbf{ }after fine-tuning.

\textbf{Loss Reduction: }Over the 50 epochs, the model’s loss steadily decreased, showing consistent improvement. It began at 2.5205 in Epoch 1 trial 0 and reduced to 0.8256 by Epoch 50, effectively minimizing error throughout training. A key outcome of this experiment was the final \textbf{13.65\% }reduction in inference time, minimizing to 0.000835 seconds per image, which underscores the efficiency of our approach compared to the baseline GoogLeNet's 0.000967 seconds. \textit{See Table 2.}

\subsection{DenseNet:}
\textbf{Performance Progression Across Epochs:} The model began with a validation accuracy of 9.66\% at Epoch 1 (Trial 0) but exhibited minimal improvement by Epoch 10, reaching 10.34\%. However, Trial 1 demonstrated significant progress, starting at 27.53\% and achieving a remarkable 81.33\% by Epoch 10. The model achieved its highest validation accuracy of 86.65\% in Trial 1, with a layer sparsity of 0.3779 (62\% of weights pruned). After fine-tuning, the quantum-inspired DenseNet achieved a test accuracy of 88.52\%, an improvement from the 86.65\%, approaching the baseline accuracy (94.05\%).

\textbf{Loss Reduction: }The model demonstrated steady loss reduction throughout the training process, beginning at 2.3028 during Epoch 1 in Trial 0 and decreasing to 0.5606 by Epoch 10 in Trial 1, indicating effective error minimization. This consistent decline reflects the model's ability to optimize its parameters and improve performance across trials. One of the standout results of this experiment was the final reduction in inference time by\textbf{ 15.20\%}, dropping to 0.001043 seconds/image, marking a considerable improvement compared to the baseline DenseNet's 0.00123 seconds. \textit{See Table 3.}

\subsection{ResNet-18:}
\textbf{Performance Progression}: Throughout the trials, there were notable fluctuations in performance. The highest validation accuracy across all trials peaked at 91.42\% in Trial 4, where the model achieved a layer sparsity of approximately 0.3779 (meaning nearly 62\% of the weights were pruned while maintaining performance). After extensive fine-tuning, the final quantum-inspired ResNet-18 reached a test accuracy of 87.11\%, a significant improvement from the earlier 84.56\% (the highest accuracy before the final fine-tuning) and approaching the baseline accuracy of 93.11\%.

\textbf{Loss Reduction}: The loss reduction across trials also followed a clear downward trend. In Trial 3, with 0.4805 layer sparsity and a rank of 10, the validation loss dropped from 1.9847 to 0.6321 (first to tenth epoch), indicating better model convergence. Notably, inference time dropped by 36.4\% to 0.00007 seconds/image, improving from the baseline of 0.00011 seconds/image \textit{See Table 4.}

%%%%%%%%%%%%%%%%%%%%%%%%%%%%%%%%%%%%%%%%%%%%%%%%%%%%%%%%%%%%

%%%%%%%%%%%%%%%%%%%%%%%%%%%%%%%%%%%%%%%%%%%%%%%%%%%%%%%%%%%%

\newpage
\section*{NeurIPS Paper Checklist}

\begin{enumerate}

\item {\bf Claims}
    \item[] Question: Do the main claims made in the abstract and introduction accurately reflect the paper's contributions and scope?
    \item[] Answer: \answerYes{} 
    \item[] Justification: The introduction clearly introduces QIANets and how the method aims to integrate the quantum-inspired strategies, accurately reflecting the paper's scope. The claims made in the abstract/introduction are supported by the results obtained.

\item {\bf Limitations}
    \item[] Question: Does the paper discuss the limitations of the work performed by the authors?
    \item[] Answer: \answerYes{}
    \item[] Justification: The paper discusses limitations in section 6, explaining the data and hardware constraints and limited number of runs. We state the lack of adaptation and provide avenues for future work to experiment on a more expansive range of models and datasets.

\item {\bf Theory Assumptions and Proofs}
    \item[] Question: For each theoretical result, does the paper provide the full set of assumptions and a complete (and correct) proof?
    \item[] Answer: \answerNA{}
    \item[] Justification: The paper does not include any theoretical results.

    \item {\bf Experimental Result Reproducibility}
    \item[] Question: Does the paper fully disclose all the information needed to reproduce the main experimental results of the paper to the extent that it affects the main claims and/or conclusions of the paper (regardless of whether the code and data are provided or not)?
    \item[] Answer: \answerYes{}
    \item[] Justification: The code and instructions on how to reproduce the experiments are included in the GitHub link. Within the paper, the model is described in detail with code snippets and is reproducible using the information provided.

\item {\bf Open access to data and code}
    \item[] Question: Does the paper provide open access to the data and code, with sufficient instructions to faithfully reproduce the main experimental results, as described in supplemental material?
    \item[] Answer: \answerYes{}
    \item[] Justification: Code is provided with instructions through the GitHub link.

\item {\bf Experimental Setting/Details}
    \item[] Question: Does the paper specify all the training and test details (e.g., data splits, hyperparameters, how they were chosen, type of optimizer, etc.) necessary to understand the results?
    \item[] Answer: \answerYes{}
    \item[] Justification: The paper covers the experimental set up, including training and test details in section 4.1. Full details are included in the code.

\item {\bf Experiment Statistical Significance}
    \item[] Question: Does the paper report error bars suitably and correctly defined or other appropriate information about the statistical significance of the experiments?
    \item[] Answer: \answerNo{}
    \item[] Justification: Statistical significance tests are not included as we could not conduct multiple trails of our finalized method due to the computational expense. As a result, error bars and confidence intervals are not included in the results. 

\item {\bf Experiments Compute Resources}
    \item[] Question: For each experiment, does the paper provide sufficient information on the computer resources (type of compute workers, memory, time of execution) needed to reproduce the experiments?
    \item[] Answer: \answerYes{}
    \item[] Justification: The paper specifies our use of an NVIDIA A40 GPU and the approximate amount of compute used in section 4.1.
    
\item {\bf Code Of Ethics}
    \item[] Question: Does the research conducted in the paper conform, in every respect, with the NeurIPS Code of Ethics \url{https://neurips.cc/public/EthicsGuidelines}?
    \item[] Answer: \answerYes{}
    \item[] Justification: The research conducted complies fully with all of the ethical guidelines listed.

\item {\bf Broader Impacts}
    \item[] Question: Does the paper discuss both potential positive societal impacts and negative societal impacts of the work performed?
    \item[] Answer: \answerNA{}
    \item[] Justification: The research has no potential societal impacts besides increasing the speed of responses from preexisting models.
    
\item {\bf Safeguards}
    \item[] Question: Does the paper describe safeguards that have been put in place for responsible release of data or models that have a high risk for misuse (e.g., pretrained language models, image generators, or scraped datasets)?
    \item[] Answer: \answerNA{}
    \item[] Justification: The paper does not introduce/include data or models with such risks. 

\item {\bf Licenses for existing assets}
    \item[] Question: Are the creators or original owners of assets (e.g., code, data, models), used in the paper, properly credited and are the license and terms of use explicitly mentioned and properly respected?
    \item[] Answer: \answerYes{}
    \item[] Justification: The paper properly cites the CIFAR-10 dataset that the models were tested on.

\item {\bf New Assets}
    \item[] Question: Are new assets introduced in the paper well documented and is the documentation provided alongside the assets?
    \item[] Answer: \answerYes{} 
    \item[] Justification: All the code is provided as supplemental material, and there are no new assets besides the proposed method.

\item {\bf Crowdsourcing and Research with Human Subjects}
    \item[] Question: For crowdsourcing experiments and research with human subjects, does the paper include the full text of instructions given to participants and screenshots, if applicable, as well as details about compensation (if any)? 
    \item[] Answer: \answerNA{}
    \item[] Justification: The paper does not involve crowdsourcing nor research with human subjects.

\item {\bf Institutional Review Board (IRB) Approvals or Equivalent for Research with Human Subjects}
    \item[] Question: Does the paper describe potential risks incurred by study participants, whether such risks were disclosed to the subjects, and whether Institutional Review Board (IRB) approvals (or an equivalent approval/review based on the requirements of your country or institution) were obtained?
    \item[] Answer: \answerNA{}
    \item[] Justification: The paper does not involve crowdsourcing nor research with human subjects.

\end{enumerate}

\end{document}